\documentclass[letterpaper]{article} 
\usepackage{aaai2026}  
\usepackage{times}  
\usepackage{helvet}  
\usepackage{courier}  
\usepackage[hyphens]{url}  
\usepackage{graphicx} 
\usepackage{natbib}  
\usepackage{caption} 
\usepackage{algorithm}
\usepackage{algorithmic}

\usepackage{newfloat}
\usepackage{listings}
\DeclareCaptionStyle{ruled}{labelfont=normalfont,labelsep=colon,strut=off} 
\floatstyle{ruled}
\newfloat{listing}{tb}{lst}{}
\floatname{listing}{Listing}
\usepackage{auxlib}
\title{Richer Representations for Neural Algorithmic Reasoning via Auxiliary Reconstruction}
\author{
    Jiafu Huang\textsuperscript{\rm 1}, Chao Peng\textsuperscript{\rm 1,}\thanks{Correspondence to: Chao Peng, Chenyang Xu, Zhengfeng Yang (\{cpeng, cyxu, zfyang\}@sei.ecnu.edu.cn).}, Chenyang Xu\textsuperscript{\rm 1,}\footnotemark[1], Zhengfeng Yang\textsuperscript{\rm 1,}\footnotemark[1], Kecheng Cai\textsuperscript{\rm 1}, Chenhao Zhang\textsuperscript{\rm 1},  Yi Wang\textsuperscript{\rm 1}, Yiwei Gong\textsuperscript{\rm 1}, Wanqin Zhou\textsuperscript{\rm 2}, Irene Zheng\textsuperscript{\rm 3}\\
}
\affiliations{
    \textsuperscript{\rm 1} Shanghai Key Laboratory of Trustworthy Computing, East China Normal University, China\\

    \textsuperscript{\rm 2} Overseas Education College, Xiamen University, China\\
    \textsuperscript{\rm 3} Dept. of Electrical and Electronic Engineering, The University of Melbourne, Australia\\
}

\usepackage{bibentry}
\usepackage{booktabs}
\begin{document}

\maketitle

\begin{abstract}
Neural algorithmic reasoning has recently emerged as a popular research direction. It 
aims to train neural networks to mimic the step-by-step behavior of classical rule-based algorithms. More specifically, the execution of such algorithms can be abstracted as a sequence of states, where each state represents the intermediate outcome after an execution step. The training objective is to generate state sequences that replicate the underlying algorithmic process. A common framework for this task adopts an ``encoder-processor-decoder'' architecture, where the encoder learns representations of states, the processor simulates algorithmic steps, and the decoder reconstructs output states. While prior work has primarily focused on improving the processor, the role of the encoder in representation learning has received little attention. Most existing methods rely on simple MLP encoders, raising the question of whether such representations are sufficiently informative for supporting algorithmic reasoning.

This paper investigates how to improve encoder representations for neural algorithmic reasoning. We propose a reconstruction module that aims to recover the input state from its encoded representation. This auxiliary reconstruction task encourages the encoder to retain critical information about the input. We demonstrate that incorporating this task during training improves the performance of existing neural architectures on standard benchmarks. Furthermore, we observe that current encoders often underutilize the correlations among features within a state. To address this, we draw inspiration from self-supervised learning and design an enhanced variant of the auxiliary task that encourages the encoder to capture intra-state feature dependencies. Experimental results show that our method enables the encoder to learn richer representations, thereby enhancing the performance of existing processors on algorithmic reasoning tasks.

\end{abstract}

\begin{links}
\end{links}

\section{Introduction}

Neural algorithmic reasoning (NAR), which bridges the gap between discrete algorithmic logic and continuous neural representations, has received growing attention in recent years. The central goal of this direction is to enable neural networks to perform algorithmic reasoning. In particular, the execution of a classical rule-based algorithm can be abstracted as a sequence of states (see \Cref{sec:state} for more details), and NAR seeks to train neural models to generate similar state sequences given an initial input, thereby mimicking the step-by-step behavior of classical algorithms. This ability to internalize algorithmic structure allows neural networks to serve as algorithmically informed components within larger architectures, supporting tasks such as algorithm-guided planning~\cite{DBLP:conf/nips/DeacVMBTN21}, neural program synthesis~\cite{DBLP:conf/iclr/NumerosoBV23}, and generalizable reasoning over structured inputs such as graphs and sequences~\cite{DBLP:conf/log/VelickovicBKLHP22}.


This line of research was first introduced in~\cite{DBLP:conf/icml/VelickovicBBPBD22}, where the authors proposed the now-standard \emph{encoder–processor–decoder} paradigm for solving NAR tasks. In this framework, given an algorithmic input, the encoder transforms the current state into latent representations, the processor then simulates the algorithm’s transition dynamics over time, and the decoder maps the processed features into predictions of the next state. This modular architecture enables flexible and scalable learning across a wide range of combinatorial tasks.

Building on this paradigm, many follow-up studies have proposed new network architectures to enhance the model’s reasoning capabilities. Examples include TripletGMPNN~\cite{DBLP:conf/log/IbarzKPNBCDBVRD22}, RNAR~\cite{DBLP:journals/corr/abs-2409-07154}, and G-ForgetNet~\cite{DBLP:conf/iclr/BohdeLSJ24}, which each introduce structural innovations to improve performance on NAR tasks.
However, we observe that most of these efforts focus on improving the processor component, while the encoder has received comparatively little attention. In most existing methods, the encoder is implemented as simple linear layers, with limited consideration of the structural properties inherent to NAR tasks. 
This raises the following question:
\begin{center}
\begin{minipage}{0.95\linewidth}
\centering
\emph{Can we design NAR-specific encoder architectures that yield richer representations?}
\end{minipage}
\end{center}

Another limitation of the current paradigm lies in its training objective, which relies solely on the final prediction loss. Consequently, it remains unclear whether suboptimal reasoning performance stems from an underperforming processor or an encoder that fails to generate sufficiently expressive representations.
This leads to another key question:
\begin{center}
\begin{minipage}{0.95\linewidth}
\centering
\emph{Can we develop a more targeted training objective that guides the encoder independently of the processor?}
\end{minipage}
\end{center}



\subsection{Our Contributions}

The two questions raised above motivate our work. We address them by proposing a training objective specifically tailored for the encoder, along with a NAR-specific encoder architecture designed to capture the structural properties of algorithmic states, thereby enriching the learned representations and enhancing the model’s reasoning capability.


To evaluate the effectiveness of our approach, we conduct experiments on the CLRS benchmark~\cite{DBLP:conf/icml/VelickovicBBPBD22}, a widely adopted benchmark for NAR that presents two key challenges:
(1) \textbf{Diversity}: The benchmark comprises 30 diverse algorithmic tasks drawn from the classical textbook \emph{Introduction to Algorithms}~\cite{DBLP:books/daglib/0023376}, covering topics such as sorting, searching, dynamic programming, graph traversal, string processing, and more. This diversity makes it highly challenging to design a single architecture that generalizes well across tasks with fundamentally different computational structures.
(2) \textbf{Generalization}: A key design feature of CLRS is the significant scale gap between the training and test instances. While models are trained on small problem instances, they are evaluated on substantially larger ones, making generalization to unseen scales a critical challenge. Across most tasks, the training set instances contain approximately 10 input elements on average (e.g., list items or graph nodes), whereas the test instances contain up to 64 elements. This creates a substantial difficulty, as models must generalize from small training inputs to much larger and more complex test cases.

Our contributions are summarized as follows:
\begin{itemize}
    \item We augment the standard encoder–processor–decoder framework with an \emph{ additional reconstruction module} and propose an \emph{auxiliary reconstruction objective} that provides direct supervision to the encoder. This objective encourages the encoder to retain essential information about the current algorithmic state, independent of the downstream processor.

    \item Built on the extended framework, we design a more expressive encoder architecture tailored to NAR tasks. In contrast to simple linear projections, our encoder incorporates a graph neural network and gated residual connections to better capture the algorithmic structures. This design enables the encoder to effectively support both the prediction and reconstruction tasks simultaneously.

    \item We further introduce a feature-level masking strategy that enhances representation learning within the proposed framework. This strategy is inspired by self-supervised learning and explicitly leverages the correlations between different components of a NAR state, improving the encoder’s ability to capture intra-state dependencies.

    \item We conduct comprehensive experiments on the challenging CLRS benchmark, demonstrating that our method yields consistent improvements across a wide range of algorithmic tasks. ReNAR significantly boosts the average accuracy over baseline models, improving it from 83.63\% to 87.11\%. M-ReNAR achieves further gains by leveraging the proposed masking strategy, reaching an average accuracy of 88.41\%.

\end{itemize}

\subsection{Other Related Works}

\paragraph{Representation Learning.}  
Representation learning aims to discover meaningful features from raw data, facilitating downstream tasks without the need for manual feature engineering. It has been a central focus across a wide range of domains, including computer vision~\cite{DBLP:conf/cvpr/HeZRS16, DBLP:conf/iclr/DosovitskiyB0WZ21}, natural language processing~\cite{DBLP:journals/corr/abs-1802-05365,DBLP:conf/naacl/DevlinCLT19}, and graph-structured data~\cite{DBLP:conf/iclr/KipfW17,DBLP:conf/iclr/HuLGZLPL20}. A variety of techniques based on self-supervised learning have been developed for effective representation learning. 
These methods typically rely on auxiliary objectives~\cite{DBLP:conf/iccv/DoerschGE15,DBLP:conf/nips/RongBXX0HH20} or structural priors~\cite{DBLP:journals/corr/abs-1806-01261} to encourage models to capture semantic or structural patterns inherent in the data.


\paragraph{Self-Supervised Learning.} 
Self-supervised learning has emerged as a powerful paradigm for learning meaningful representations without the need for large-scale labeled data. Techniques such as contrastive learning \cite{DBLP:conf/icml/ChenK0H20, DBLP:conf/cvpr/He0WXG20} and masked prediction\cite{DBLP:conf/naacl/DevlinCLT19,DBLP:conf/iclr/Bao0PW22} have achieved remarkable success across domains by encouraging models to capture internal correlations and latent semantics. Foundational efforts in graph self-supervised learning have extended these principles to structured data, proposing general objectives such as subgraph-level discrimination and attribute masking  \cite{DBLP:conf/nips/YouCSCWS20,DBLP:conf/iclr/HuLGZLPL20}, which have proven effective for learning transferable representations. Building on these advances, graph autoencoders, such as SimGRACE\cite{DBLP:conf/www/XiaWCHL22} and GraphMAE\cite{DBLP:conf/kdd/HouLCDYW022} has emerged, leveraging contrastive or masked reconstruction strategies to further enhance the quality of graph representations. 

\section{Preliminaries}\label{sec:pre}


In this section, we formally introduce the setting of neural algorithmic reasoning and describe the encoder-processor-decoder paradigm commonly used in the literature.


\subsection{Classical Algorithm Execution as State Sequences}\label{sec:state}

We begin by discussing how classical rule-based algorithms can be abstracted as sequences of structured states. These algorithms typically maintain internal variables that evolve over time, and the values of these variables at each step can be interpreted as discrete execution states.


\begin{figure}[tb]
    \centering
    \includegraphics[width=0.65\linewidth]{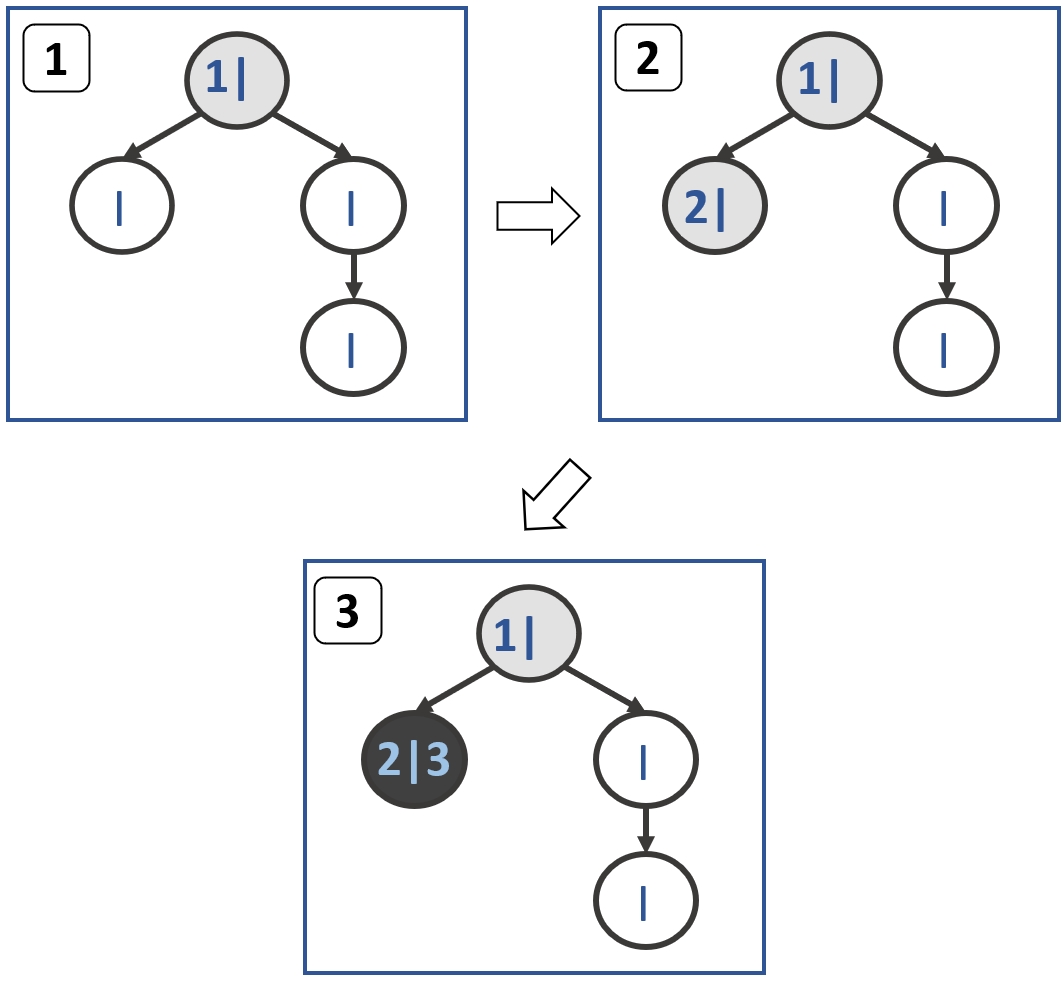}
    \caption{An illustration of three execution steps of the DFS algorithm on a sample graph. Each node is annotated with its discovery and finishing times in the format $d\,|\,f$, and the color indicates its visitation status.
   }
    \label{fig:state}
\end{figure}

To illustrate this idea, consider the depth-first search (DFS) algorithm~\cite{DBLP:books/daglib/0023376}. At each step, DFS maintains a state composed of node-level information\footnote{We remark that DFS also maintains certain edge-level variables, such as whether an edge has been explored. However, for simplicity, we focus here solely on node-level information to provide a clean illustration.} including \emph{color}, \emph{discovery time} $d$, and \emph{finishing time} $f$.
The color of a node indicates its visitation status in the traversal: white for unvisited, gray for discovered but not fully explored, and black for fully explored. The discovery time $d$ records the time when a node is first encountered, and the finishing time $f$ marks the time when the exploration of all its descendants is complete.

\Cref{fig:state} illustrates three execution steps of DFS on a given instance. Each state can be viewed as a structured snapshot of the graph, annotated with the corresponding node-level variables. In NAR benchmarks, these variables are typically referred to as \emph{hints}~\cite{DBLP:conf/icml/VelickovicBBPBD22}. 

The collection of all (node, hint) pairs at a specific time step constitutes the algorithm's state at that point. The sequence of such states over time forms the complete execution trajectory of the algorithm on the input instance.

\subsection{NAR Datasets}

A typical NAR dataset is defined with respect to a specific algorithmic problem and the classical algorithm chosen to solve it. Each instance in the dataset consists of a problem input (usually represented as a graph) along with its corresponding execution trace, encoded as a sequence of states. 

Formally, let $\vx$ denote a problem instance (e.g., a graph with node and edge features), and let $\vy = \{\vy^{(1)}, \vy^{(2)}, \ldots, \vy^{(T)}\}$ denote the corresponding execution trajectory. Each $\vy^{(t)}$ represents the state of the algorithm at time step $t$, consisting of a set of hints defined over the nodes and edges of the graph. This sequence serves as the target output for neural networks trained to emulate the algorithm in a step-by-step manner.

\subsection{Encoder-Processor-Decoder Paradigm}\label{sec:para}

The \emph{encoder–processor–decoder} paradigm serves as the standard framework in most prior work. At each reasoning step, the model receives as input the pair $\left( \vx, \vy^{(t)} \right)$, where $\vx$ is the problem instance and $\vy^{(t)}$ is the current state. The goal is to predict the next state $\vy^{(t+1)}$.

The encoder module consists of multiple encoder networks. Suppose there are $k$ distinct hint types; accordingly, the encoder module includes $k$ separate encoder networks, each responsible for processing one specific hint. Let $G = (V, E)$ denote the input graph structure. The state at time step $t$ is represented as $\vy^{(t)} = \{ \vh_1^{(t)}, \ldots, \vh_k^{(t)} \}$, where each $\vh_i^{(t)} = \{ h_{i}^{(t)}(v) \}_{v \in V}$ denotes the value of the $i$-th hint over all nodes in the graph\footnote{For simplicity, we describe only node-level hints in this illustration and in the descriptions of the following sections. The benchmark datasets may also include edge-level and graph-level hints, which can be handled in a similar manner.}.

Each encoder network takes as input the problem instance $\vx$ and a specific hint $\vh_i^{(t)}$, and produces a corresponding representation $\vz_i^{(t)} = \{ z_{i}^{(t)}(v) \}_{v \in V}$ for all nodes. These hint-wise representations are then aggregated and passed to the processor module, which is typically implemented using a graph neural network. The processor simulates the algorithm's transition dynamics and generates latent node features. These features are then fed into the decoder module, which also consists of $k$ decoder networks. Each decoder is responsible for predicting one specific hint, and together they produce the predicted output state $\vhy^{(t+1)}=\{\vhh^{(t+1)}_i\}_{i\in [k]}$.
To supervise the model's predictions, we compare the predicted state $\vhy^{(t+1)}$ with the ground-truth state $\vy^{(t+1)}$ at each time step and compute a prediction loss over all hint types. Let $\mathcal{L}_{\text{pred}}^{(t)}$ denote the prediction loss at step $t$, defined as $\mathcal{L}_{\text{pred}}^{(t)} = \sum_{i=1}^{k} \ell_{\text{hint}}\left( \vhh_i^{(t+1)}, \vh_i^{(t+1)} \right),$
where $\vhh_i^{(t+1)}$ and $\vh_i^{(t+1)}$ denote the predicted and ground-truth values of the $i$-th hint, respectively. The function $\ell_{\text{hint}}$ is a hint-specific loss function, chosen based on the type of hint (e.g., cross-entropy for classification-type hints like node color in DFS; mean squared error for regression-type hints like discovery or finishing times).
The model is trained by minimizing the total prediction loss across all time steps and training examples. All components of the encoder–processor–decoder architecture are optimized jointly using backpropagation.

\section{Incorporating Auxiliary Reconstruction}\label{sec:re}



As described in the previous section, standard neural algorithmic reasoning models are trained to minimize a prediction loss between the predicted state and the ground-truth state. While this loss effectively guides the overall training, it entangles the contributions of the encoder, processor, and decoder, making it difficult to isolate the quality of the encoder's representations. In particular, when the model underperforms, it remains unclear whether the issue lies in the processor's transition modeling or in the encoder's ability to produce informative representations.

To address this ambiguity and to explicitly encourage high-quality representation learning at the encoder stage, we introduce an auxiliary reconstruction module. Reconstruction is a widely used self-supervised objective in representation learning~\cite{DBLP:conf/icml/VincentLBM08,DBLP:journals/corr/KingmaW13,DBLP:journals/corr/KipfW16a}, where the goal is to recover the input from its latent representation. By applying this principle, we design an additional loss that directly supervises the encoder output, encouraging it to retain sufficient information about the input state. This auxiliary task complements the main prediction loss and provides a more targeted signal for training the encoder effectively.

\subsection{Framework with Reconstruction Module}

We now describe the extended framework that incorporates an auxiliary reconstruction module, which provides additional supervision to the encoder by encouraging it to retain information sufficient to recover the input state.

\begin{figure}[tb]
    \centering
    \includegraphics[width=0.95\linewidth]{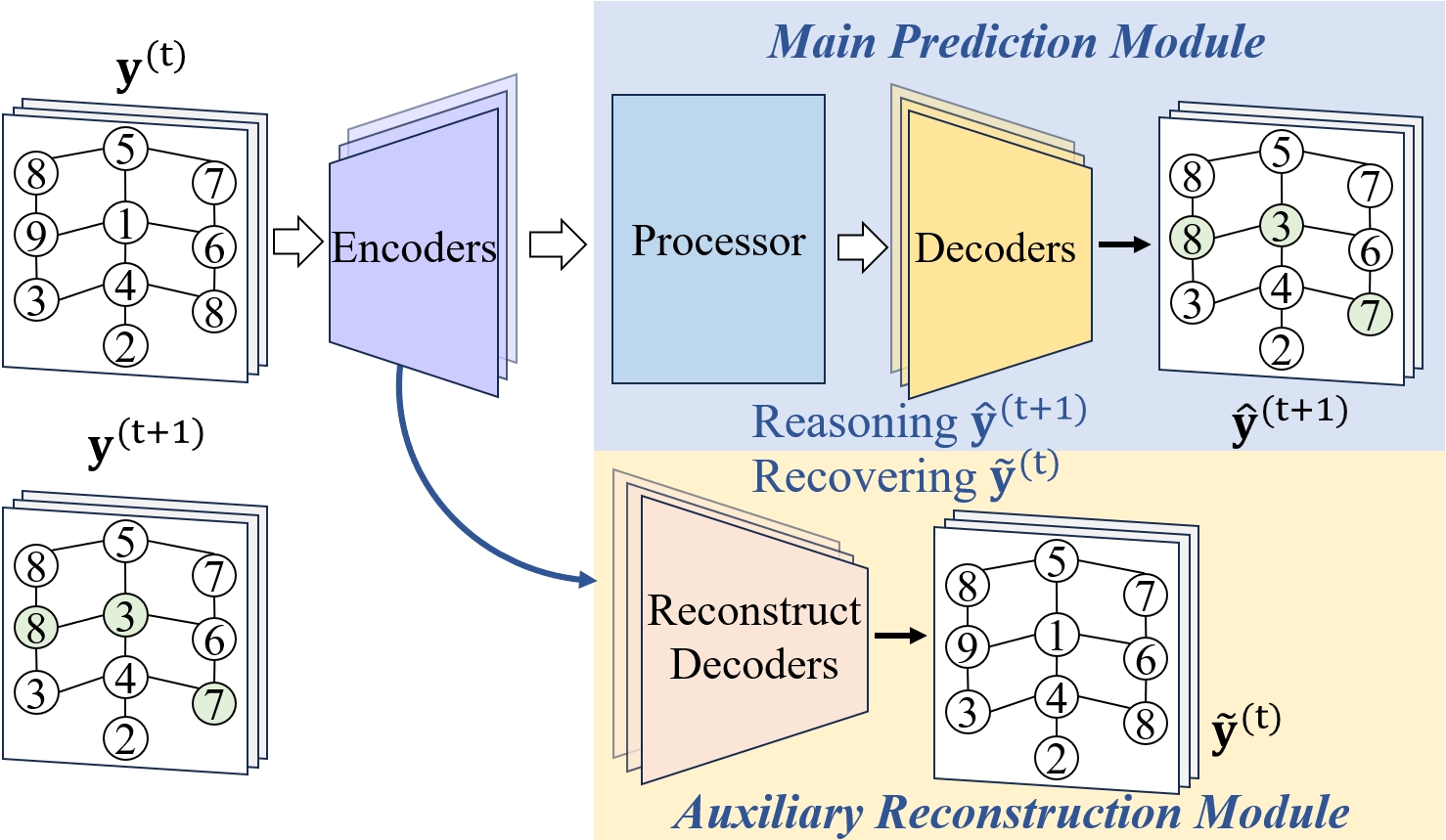}
    \caption{An extended encoder–processor–decoder framework incorporating an auxiliary reconstruction module. }
    \label{fig:framework}
\end{figure}

An overview of the framework is shown in \Cref{fig:framework}. The reconstruction module branches from the encoder and operates in parallel with the main prediction pathway. In addition to predicting the next algorithmic state, the encoder outputs are also used to reconstruct the current input state, thereby providing an additional learning signal.

More precisely, the reconstruction module consists of $k$ reconstruction decoders, with the same structure and number as the original decoders. Each reconstruction decoder is responsible for reconstructing one specific hint, similar to how each original decoder predicts a hint in the next state. However, unlike the original decoders, which take the processor’s latent output to predict the next state \(\vhy^{(t+1)} = \{ \vhh^{(t+1)}_i \}_{i \in [k]}\), the reconstruction decoders take as input the encoder representations \(\vz^{(t)}\) and attempt to reconstruct the current state. That is, they produce \(\vty^{(t)} = \{ \vth^{(t)}_i \}_{i \in [k]}\), where each \(\vth^{(t)}_i\) is the reconstruction of the hint \(\vh^{(t)}_i\) from input state \(\vy^{(t)}\) at time step \(t\).

To supervise the reconstruction branch, we define a reconstruction loss that compares the reconstructed state \(\vty^{(t)}\) with the input state \(\vy^{(t)}\). Similar to the prediction loss, the reconstruction loss is computed over all hint types $\mathcal{L}_{\text{rec}}^{(t)} =\sum_{i=1}^{k} \ell_{\text{hint}}\left( \vth_i^{(t)}, \vh_i^{(t)} \right),$
where the same hint-specific loss function \(\ell_{\text{hint}}\) used in the prediction loss is applied here.

To jointly optimize both the prediction and reconstruction objectives, we combine the two losses at each time step with a weighting coefficient \(\lambda\). The total loss is defined as $\mathcal{L}_{\text{total}} = \sum_{t=1}^{T-1} \left( \mathcal{L}_{\text{pred}}^{(t)} + \lambda \cdot \mathcal{L}_{\text{rec}}^{(t)} \right).$


This training objective supervises both the main prediction path and the auxiliary reconstruction path, encouraging the encoder to produce representations that are not only useful for predicting future states but also sufficiently informative to recover the current state.


\subsection{Reconstruction-Oriented Encoder Design}

In the framework introduced above, the auxiliary reconstruction module places additional demands on the encoder: it must not only support accurate prediction of future states through the processor but also preserve sufficient information to reconstruct the current state directly. To meet this dual objective, we revisit the design of the encoder networks.

In previous NAR methods, each encoder network is typically implemented as a simple linear layer applied to the corresponding hint.
However, this design\footnote{Additional empirical evidence is provided in \cref{sec:abl1}
} remains limited in its ability to capture structural context. To further enhance the encoder's expressive power, we augment encoder networks with a graph neural network layer and a gated network. These modifications are designed to strengthen the encoder's capacity to model local structure and retain key information relevant for reconstruction.

\begin{figure}[tb]
    \centering
    \includegraphics[width=0.99\linewidth]{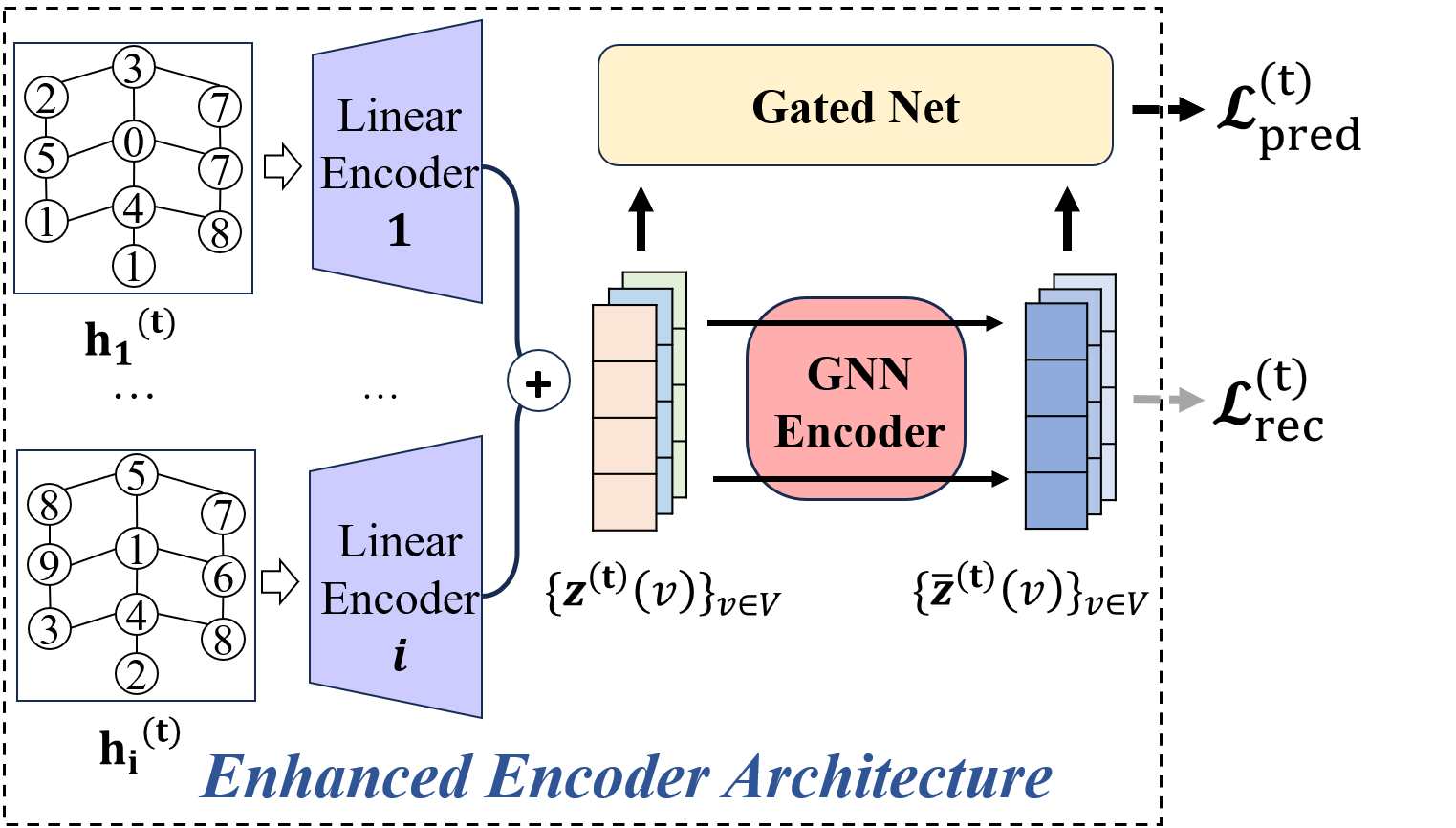}
    \caption{An illustration of the enhanced encoder. }
    \label{fig:encoder}
\end{figure}

An illustration of the new encoder module is shown in~\cref{fig:encoder}. Each hint $\vh_i^{(t)}$ is first passed through its corresponding linear layer, producing node-level latent features $\{ z_i^{(t)}(v) \}_{v \in V}$ for the $i$-th hint. These features are then aggregated across hints for each node to form the initial embedding for the GNN encoder:
$ 
\vz^{(t)}(v) = \sum_{i \in [k]} z_i^{(t)}(v).
 $
The GNN then computes updated node representations $\vbz^{(t)}(v)$, which are used as input to the reconstruction decoders.

For the processor, we further refine the representation by combining $\vbz^{(t)}(v)$ with the initial embedding $\vz^{(t)}(v)$ through a gated mechanism:
\[
\vz^{(t)}(v) = g_v \odot \vbz^{(t)}(v) + (1 - g_v) \odot \vz^{(t)}(v),
\]
where $g_v \in [0, 1]$ is a learned gating coefficient for node $v$, and $\odot$ denotes element-wise multiplication.


Intuitively, incorporating a GNN into the encoder allows each node to aggregate local structural information at an earlier stage, enabling the encoder to generate more context-aware representations before interacting with the processor. Although the processor itself is also typically implemented as a GNN, its role is to simulate the algorithm's transition dynamics over time. In contrast, the encoder's role is to extract rich, informative features from the current state alone. 
The addition of a gated network helps preserve the original hint-specific signals, such as the raw input values for each node, while also incorporating contextual information aggregated from neighboring nodes. 
These enhancements help the encoder provide a stronger input to the processor, ultimately leading to better performance. 




\section{Advanced Reconstruction via Hint-Level Masking}\label{sec:mask}

The reconstruction framework introduced in the previous section supervises the new encoder module, which is shown in ~\cref{fig:encoder}, by requiring it to reconstruct all node-level hints in the current state. While this approach encourages the new encoder to preserve essential information, it, like previous encoder designs, treats each hint independently and does not explicitly model their interactions. In practice, however, many algorithmic states exhibit strong internal correlations across hints. Exploiting these correlations can provide additional inductive bias to help the encoder learn more structured and generalizable representations.

\subsection{Hint Dependencies}

Algorithmic reasoning tasks often involve multiple node-level hints that are tightly coupled due to the underlying algorithmic semantics. Capturing these correlations is important for learning coherent representations. 

For example, consider the DFS algorithm shown in~\cref{fig:state}, where each node is annotated with three hints: color, discovery time, and finish time. These hints are logically interdependent:
\begin{itemize}
    \item If a node has neither discovery nor finish time, its color must be white;
    \item If the discovery time is present but the finish time is not, the node must be gray;
    \item If both timestamps are available, the color must be black.
\end{itemize}   
These relationships are not imposed by external labels, but are an intrinsic part of the algorithm's execution dynamics.
Such inter-hint dependencies are also common in other algorithms, such as shortest path (where distance and predecessor are correlated). However, existing NAR frameworks typically process each hint independently in the encoder, missing the opportunity to exploit these structural regularities.

\subsection{Hint-Level Masking Strategy}

\begin{figure}[tb]
    \centering
    \includegraphics[width=0.99\linewidth]{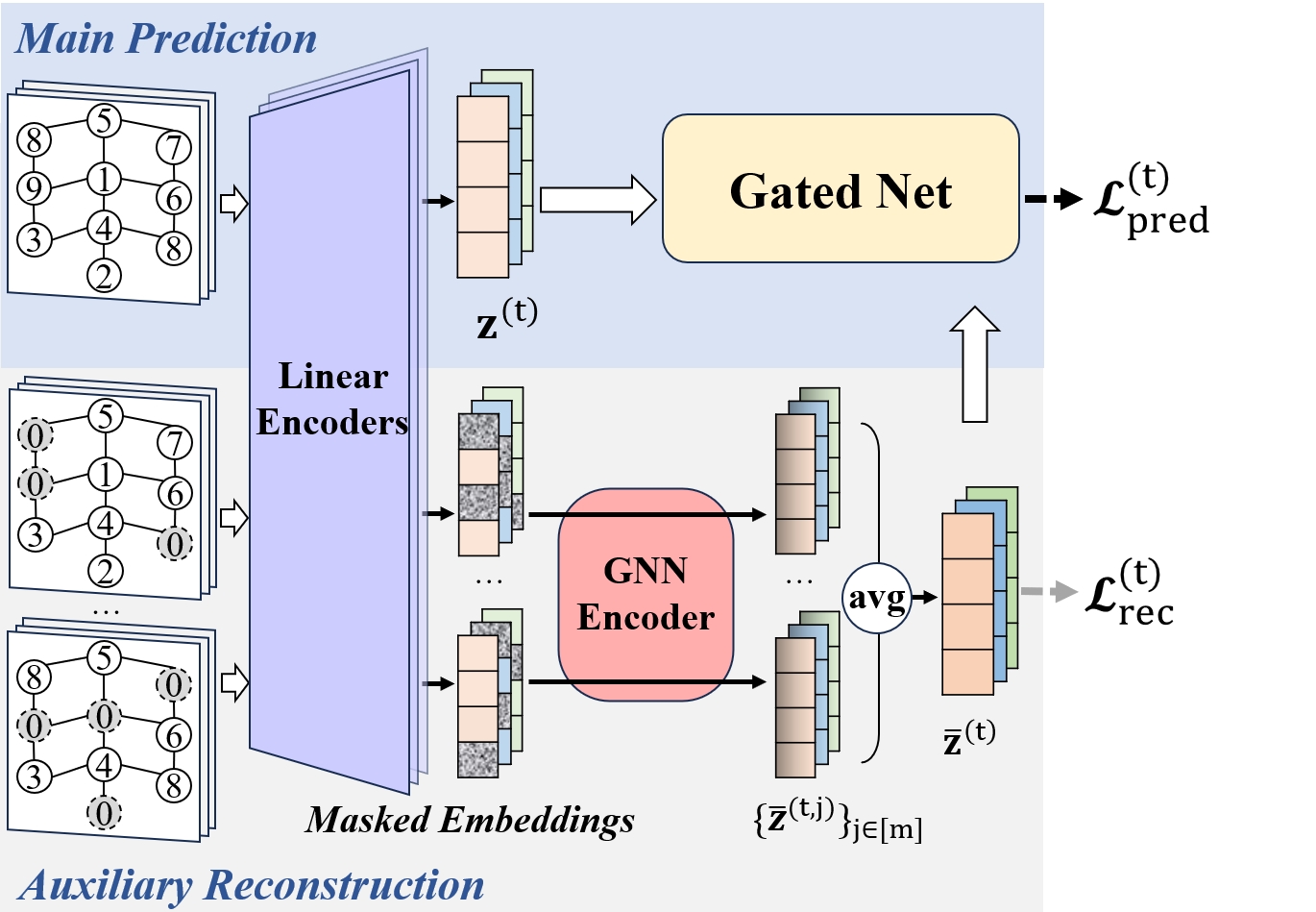}
    \caption{An illustration of the hint-level masking strategy.}
    \label{fig:mask}
\end{figure}

To encourage the new encoder to model cross-hint dependencies, we adopt a hint-level masking strategy inspired by recent advances in self-supervised learning~\cite{DBLP:conf/www/HouHCLDK023, DBLP:conf/kdd/LiWS0TZMZW23,DBLP:conf/icde/WangYHXYFZWDJ24}. Rather than treating all node-hint pairs equally during reconstruction, we selectively mask a subset of the input hints and train the encoder to recover only the masked values. This forces the encoder to infer missing information based on the context provided by the remaining unmasked hints.

Formally, we fix a masking ratio \(\beta \in (0, 1)\), which specifies the proportion of node-hint pairs to be masked in each reconstruction instance. Let \(m = \lfloor 1/\beta \rfloor\) denote the number of independent masking rounds. For each round \(j \in [m]\), we independently sample a set of masked indices \(\mathcal{M}^{(j)} \subseteq V \times [k]\), where \(|\mathcal{M}^{(j)}| = \lfloor \beta \cdot k \cdot |V| \rfloor\). 
For each \((v, i) \in \mathcal{M}^{(j)}\), the corresponding input hint \(h_i^{(t)}(v)\) is masked by setting its value to zero. The resulting masked input at round \(j\), denoted by \(\vh^{(t,j)}\), is defined element-wise as:
\[
h_i^{(t,j)}(v) = 
\begin{cases}
0, & \text{if } (v, i) \in \mathcal{M}^{(j)} \\
h_i^{(t)}(v), & \text{otherwise}.
\end{cases}
\]

The linear layers and GNN within the new encoder module, then process each masked version of the input independently, producing a collection of hidden representations \(\{ \vbz^{(t, j)} \}_{j \in [m]}\). These representations are aggregated and averaged to produce a unified representation:
$
\vbz^{(t)} = \frac{1}{m} \sum_{j=1}^{m} \vbz^{(t, j)}.
$
Subsequently, the unified representation \(\vbz^{(t)}\) is passed to the gated network along with \(\vz^{(t)}\), which is obtained by encoding the unmasked data through linear layers, to support the main prediction.

In parallel, each reconstruction decoder receives its corresponding masked input \(\vbz^{(t, j)}\) and attempts to reconstruct only the masked hints \((v, i) \in \mathcal{M}^{(j)}\). The new reconstruction loss is defined as the sum over the masked positions:
\(
\mathcal{L}_{\text{rec}}^{(t)} = \sum_{j=1}^{m} \sum_{(v, i) \in \mathcal{M}^{(j)}} \ell_{\text{hint}}\left( \tilde{h}_i^{(t,j)}(v), h_i^{(t)}(v) \right)~,
\)
where \(\tilde{h}_i^{(t,j)}(v)\) denotes the reconstruction decoder’s prediction for the node-hint pair \((v, i)\) in the \(j\)-th masking round.

\begin{figure*}[htb]
 \centering
    \includegraphics[width=0.79\linewidth]{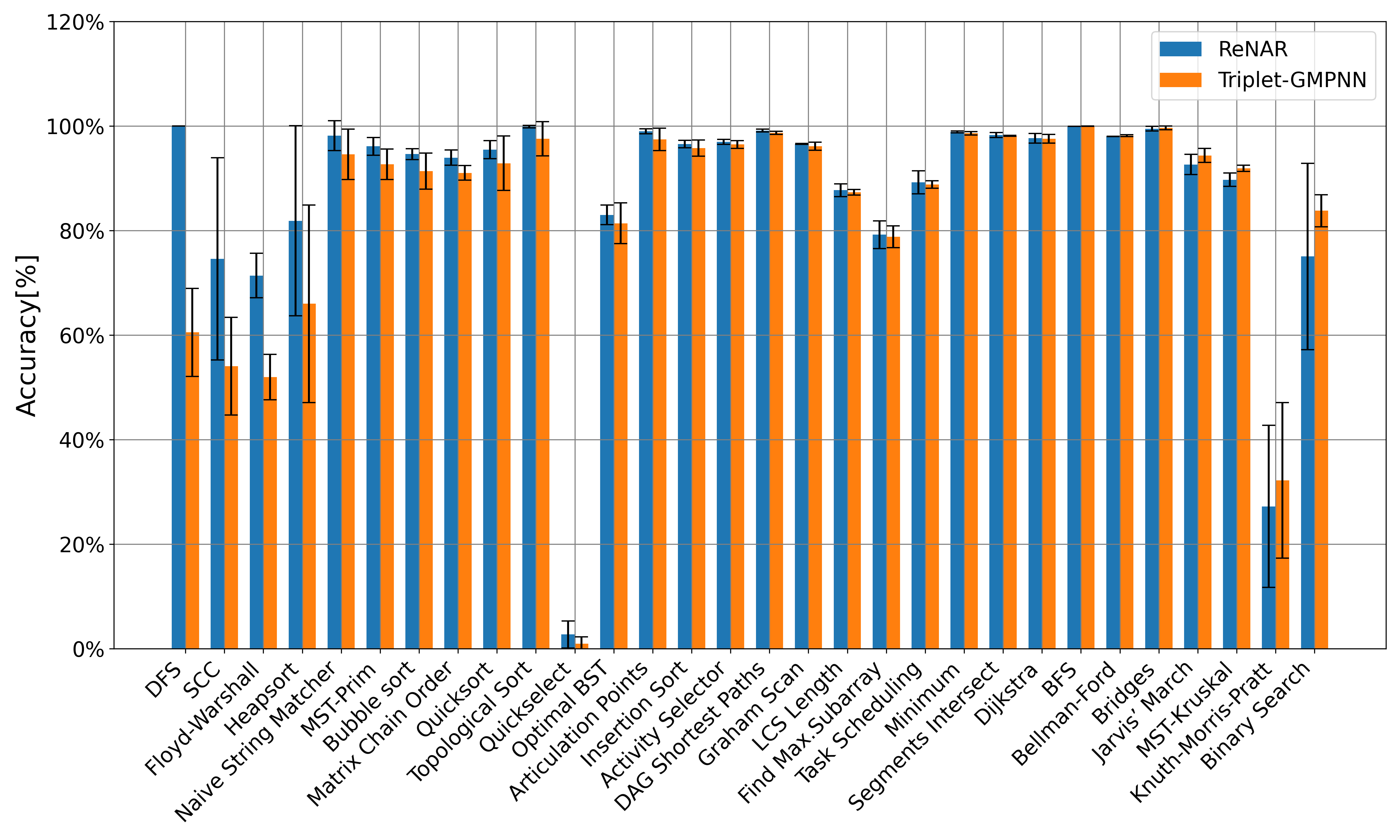}
    \caption{Comparison of the Triplet-GMPNN~\cite{DBLP:conf/log/IbarzKPNBCDBVRD22} architecture’s performance before and after augmentation with ReNAR. The 30 algorithmic tasks are arranged in descending order of the accuracy.}
    \label{fig:re_exp}
\end{figure*}

\begin{table*}[tb]
    \centering
    \small
    \captionsetup{position=bottom}
    \begin{tabular}{lcccccc}
    \toprule 
        Task Category & Prior Best & CEF-GMPNN & G-ForgetNet & Triplet-GMPNN & ReNAR & M-ReNAR \\
        \midrule 
        Graphs & 64.98\%\ $\pm$2.59 & 86.99\%\ $\pm$1.78 & 88.80\%\ $\pm$0.84 & 86.68\%\ $\pm$2.73 & \underline{93.75\%\ $\pm$2.40} & \textbf{94.74\%\ $\pm$0.73} \\
        Geometry & 92.48\%\ $\pm$1.35 & 95.38\%\ $\pm$1.11 & 95.09\%\ $\pm$1.16 & \textbf{96.21\%\ $\pm$0.74} & \underline{95.83\%\ $\pm$0.82} & 95.83\%\ $\pm$0.41 \\
        Strings & 4.08\%\ $\pm$0.57 & \underline{70.86\%\ $\pm$6.47} & 54.74\%\ $\pm$1.95 & 63.40\%\ $\pm$9.85 & 62.70\%\ $\pm$9.17 & \textbf{74.72\%\ $\pm$18.72} \\
        DP & 76.00\%\ $\pm$2.47 & 86.23\%\ $\pm$1.77 & 86.70\%\ $\pm$0.49 & 86.59\%\ $\pm$1.95 & \textbf{88.23\%\ $\pm$1.52} & \underline{87.55\%\ $\pm$1.76} \\
         Div. \& C. & 65.23\%\ $\pm$2.56 & 76.38\%\ $\pm$3.04 & 78.97\%\ $\pm$0.70 & 78.81\%\ $\pm$2.10 & \underline{79.22\%\ $\pm$2.67} & \textbf{81.54\%\ $\pm$2.71} \\
        Greedy & 84.13\%\ $\pm$2.59 & 92.61\%\ $\pm$0.44 & 91.79\%\ $\pm$0.23 & 92.65\%\ $\pm$0.73 & \underline{93.10\%\ $\pm$1.34} & \textbf{94.35\%\ $\pm$0.46} \\
        Search & 56.11\%\ $\pm$0.36 & \underline{61.74\%\ $\pm$0.45} & \textbf{63.84\%\ $\pm$0.84} & 61.14\%\ $\pm$1.57 & 58.89\%\ $\pm$6.86 & 61.22\%\ $\pm$2.04 \\
        Sorting & 71.53\%\ $\pm$0.97 & 81.52\%\ $\pm$3.18 & 78.09\%\ $\pm$3.78 & 86.52\%\ $\pm$7.28 & \textbf{92.13\%\ $\pm$5.41} & \underline{90.51\%\ $\pm$3.77} \\
        \midrule 
        Overall Average & 66.04\% & 82.68\% & 82.89\% & 83.63\%  & \underline{87.11\%}& \textbf{88.41\%} \\
    \bottomrule 
    \end{tabular}
    \caption{Summary of our results across each task category in the CLRS benchmark. The best-performing results are highlighted in bold, and the second-best results are underlined.}
    \label{tab:task_category_performance}
\end{table*}

An illustration of this masking and reconstruction process is provided in \Cref{fig:mask}. This masking strategy enables the encoder to learn more structured representations by leveraging dependencies between hints. It also introduces stochasticity into training, which we find improves generalization in downstream reasoning tasks.

\section{Experiments}\label{sec:exp}




\begin{figure}[tb]
    \centering
    \includegraphics[width=0.8\linewidth]{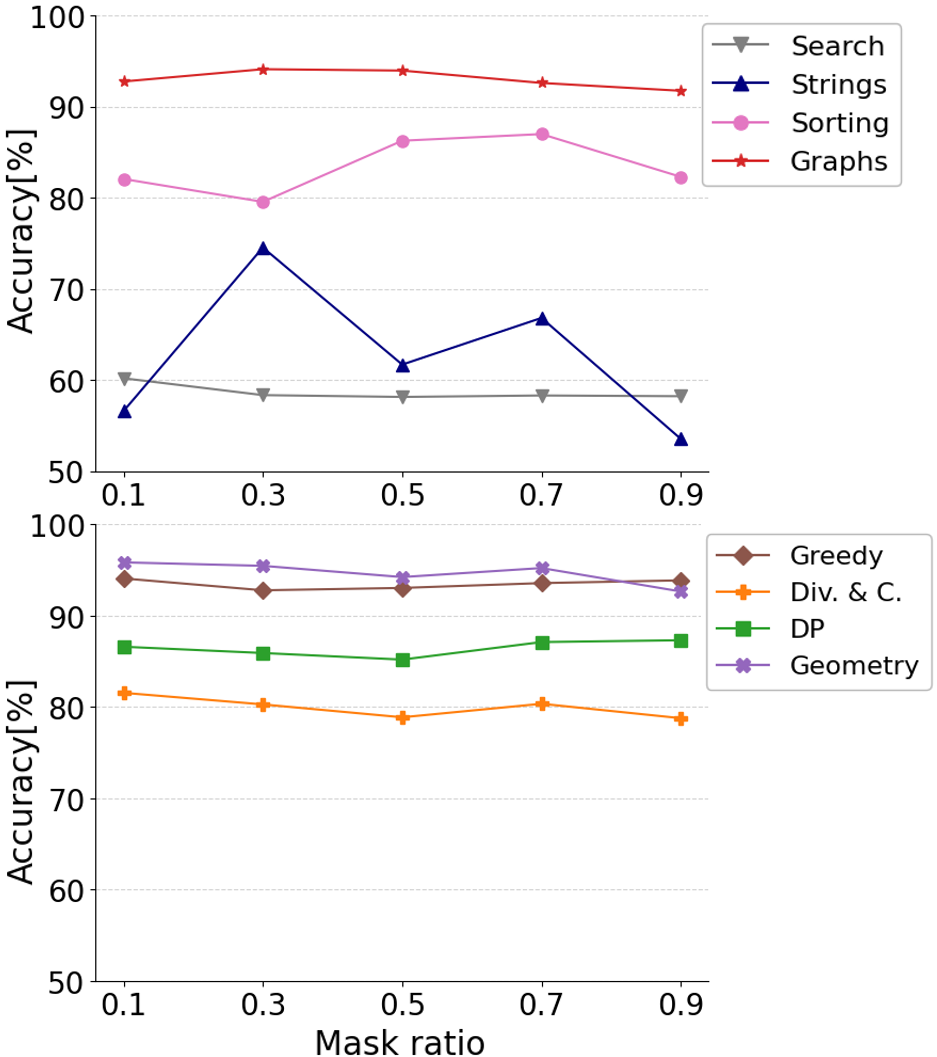}
    \caption{Performance of M-ReNAR under different masking ratios, grouped by algorithmic category. 
}
    \label{fig:maskratio}
\end{figure}

This section evaluates the effectiveness of our proposed methods on the challenging CLRS benchmark, which spans all 30 algorithmic tasks across 8 major categories. Our experiments are designed to address the following questions:

\begin{itemize}
    \item Can our framework, along with the proposed encoder architecture, be effectively integrated with existing processor modules to enhance reasoning capability?

    \item Does the hint-level masking strategy further improve performance, and how does the choice of masking ratio affect performance across different tasks?
\end{itemize}

We refer to our reconstruction-augmented framework as ReNAR\footnote{The codes are provided in https://github.com/Jeff-Huang-SHU/ReNAR.}, and the masking-enhanced variant as M-ReNAR. To investigate the first question, we integrate ReNAR with several existing processor architectures to assess whether it consistently improves reasoning performance. We report results using the widely adopted Triplet-GMPNN processor~\cite{DBLP:conf/log/IbarzKPNBCDBVRD22} in the main body, and defer the results with other processors to \cref{sec:abl3}
To answer the second question, we evaluate the performance of M-ReNAR under varying mask ratios~$\beta$. We also conduct ablation studies to isolate the contributions of individual components, with the corresponding results also provided in \cref{sec:abl2}



    


\subsection{Setup}

\paragraph{Baselines.} We compare our methods against several recent approaches, including Triplet-GMPNN~\cite{DBLP:conf/log/IbarzKPNBCDBVRD22}, CEF-GMPNN~\cite{DBLP:conf/ijcai/Shi0XY24}, and G-ForgetNet~\cite{DBLP:conf/iclr/BohdeLSJ24}. 
We also include earlier baselines such as MPNN~\cite{DBLP:conf/icml/GilmerSRVD17}, PGN~\cite{DBLP:conf/nips/VelickovicBOPVB20},  MemNet~\cite{DBLP:conf/icml/VelickovicBBPBD22}, and NPQ~\cite{npq}. For these methods, we report the best result across all algorithmic tasks and denote it as the \emph{prior best}.
We follow the standard evaluation setting on CLRS and report the F1 score as the primary metric, which is computed by averaging over all hints at each step. 

\paragraph{Computational Details.} 
Our experiments are conducted on multiple machines equipped with different NVIDIA GPUs, including RTX 4090, RTX A100, and RTX A6000. All reported results are averaged over four runs. To ensure fair comparisons, we follow the widely used experimental hyperparameter settings from~\cite{DBLP:conf/log/IbarzKPNBCDBVRD22}: the batch size is set to 32, and the models are trained for 10{,}000 steps using the Adam optimizer with a learning rate of 0.001.
For ReNAR, we set the weight of the reconstruction loss to $\lambda = 0.1$. For M-ReNAR, which includes masked reconstruction, we use a higher $\lambda = 1$ to emphasize the supervision signal. The average training time is approximately 0.79 GPU hours for ReNAR and 1.30 GPU hours for M-ReNAR, while the compared baseline method ~\cite{DBLP:conf/log/IbarzKPNBCDBVRD22} has a training time of 0.52 GPUs.

\subsection{Effectiveness of Auxiliary  Reconstruction}

This subsection evaluates the effectiveness of the proposed framework, focusing on whether it enhances the encoder’s representation quality and improves reasoning performance when integrated with existing processors. We observe that different processors exhibit consistent trends. Therefore, we report only the results based on the Triplet-GMPNN processor in the main body and provide the remaining results in \cref{sec:abl3}

As shown in~\Cref{fig:re_exp}, augmenting existing processors with our reconstruction framework leads to consistent improvements in reasoning performance across most of tasks. The gains are particularly pronounced on graph-based algorithms and sorting-related problems. For example, on the DFS reasoning task, the accuracy improves dramatically, from around 60\% to nearly 100\%.

\subsection{Impact of Hint-Level Masking}

This subsection investigates the role of the hint-level masking strategy in further improving performance and analyzes how the masking ratio~$\beta$ affects different algorithmic tasks. In our experiments, M-ReNAR is evaluated under a range of mask ratios \{0.1, 0.3, 0.5, 0.7, 0.9\}. 
We observe that tasks within the same algorithmic category tend to exhibit similar sensitivities to the masking ratio. Therefore, we compute the average accuracy for each algorithm category at different masking ratios. The results are presented in~\Cref{fig:maskratio}.

From the figure, we observe that the sensitivity to the masking ratio varies across algorithmic categories. Different algorithmic categories achieve their best performance at different values of~$\beta$. Notably, graph-based tasks are relatively stable across all settings, while string-related tasks exhibit more pronounced fluctuations. This may be due to the fact that string algorithms often rely heavily on precise token-level information, making them more sensitive to partial hint removal during training.


We summarize the comparison between our methods and the baselines in~\Cref{tab:task_category_performance}. For the M-ReNAR column, we report the best performance achieved across all tested masking ratios.
The table indicates that incorporating the reconstruction framework improves the average accuracy across the entire benchmark from 83.63\% to 87.11\%. With the addition of the hint-level masking strategy, the performance is further boosted to 88.41\%. Our methods demonstrate consistent improvements across most algorithmic categories, with particularly notable gains on graph-based tasks, where the best average accuracy improves from 88.80\% to 94.74\%.

\subsection{Discussion}

We see the following trends from the experimental results:
\begin{itemize}
    \item The proposed ReNAR framework consistently improves reasoning performance. The reconstruction objective provides direct supervision to the encoder, encouraging it to retain more task-relevant information. This, in turn, leads to stronger intermediate features that better support downstream algorithmic reasoning.

    \item The hint-level masking strategy further strengthens the reconstruction framework by enabling the encoder to capture correlations among different hints. This leads to richer representations and results in additional performance gains on the benchmark.
\end{itemize}

\section{Conclusion}\label{sec:con}

This paper introduces ReNAR, a reconstruction-augmented framework for neural algorithmic reasoning, along with a tailored encoder architecture and a hint-level masking strategy to enhance representation learning. Experiments demonstrate that our method consistently improves reasoning performance across diverse algorithmic tasks, highlighting the importance of direct encoder supervision and intra-state feature modeling.
There remain many directions for future work. For example, exploring other forms of auxiliary supervision, like contrastive or predictive objectives, may further enhance representation quality. 


\section*{Acknowledgements}
This work is supported by the National Key Research and Development Program of China (2023YFA1009402, 2025YFC2423000), NSFC Programs (62302166, 62161146001, 62372176), Shanghai Key Lab of Trustworthy Computing, Shanghai Frontiers Science Center of Molecule Intelligent Syntheses and Fundamental Research Funds for the Central Universities.
\bibliography{ref}

\appendix
\onecolumn
\newpage
\section{Ablation: Necessity of the Encoder Design}\label{sec:abl1}


To assess the necessity of the proposed encoder design, we investigate a variant of ReNAR, referred to as ReNAR-linear, in which the encoder is implemented using simple linear layers without any GNN-based enhancements. This variant is trained with the same auxiliary reconstruction objective as the full ReNAR model.

\begin{figure}[htb]
    \centering
    \includegraphics[width=0.9\linewidth]{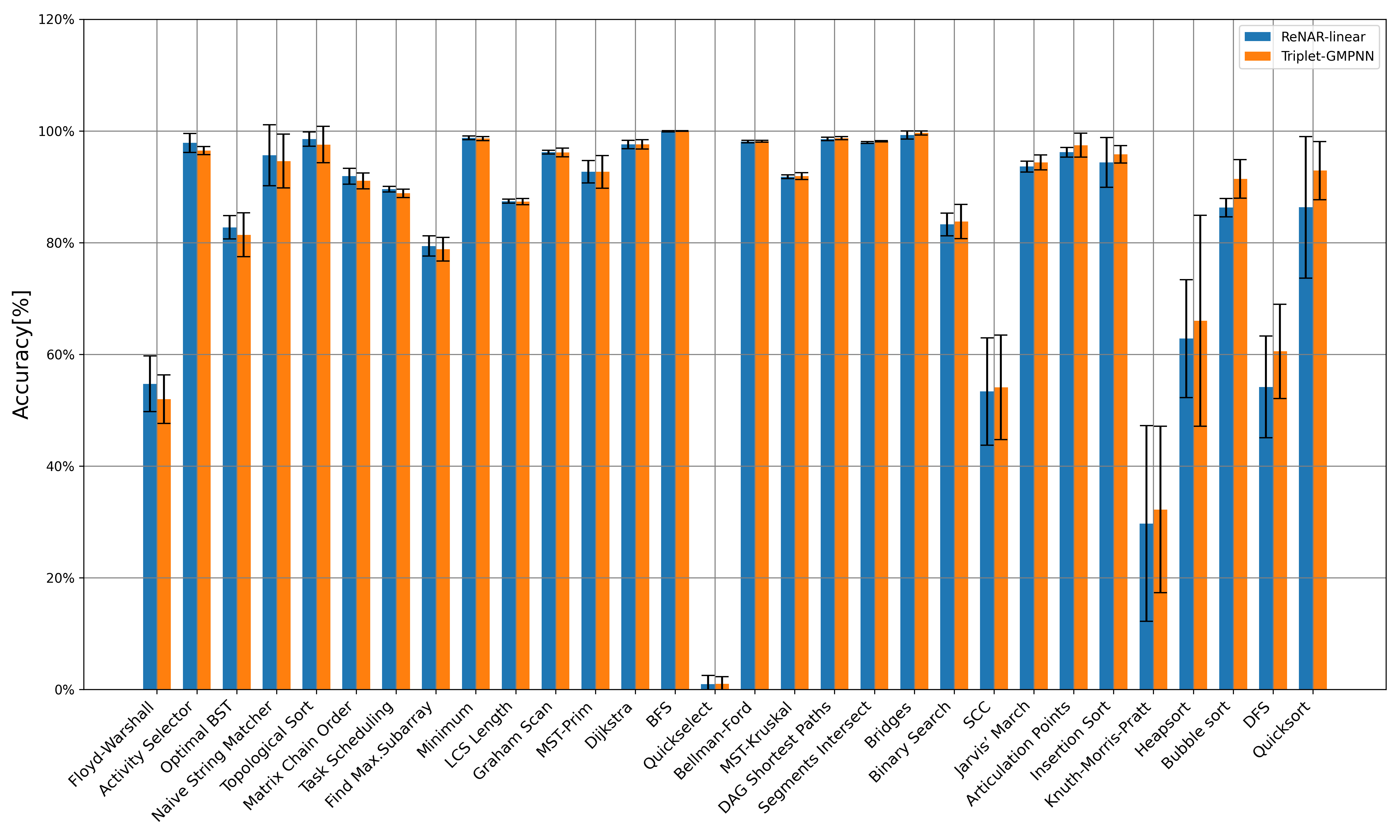}
    \caption{Comparison of the Triplet-GMPNN architecture’s performance before and after augmentation with ReNAR-linear. The 30 algorithmic tasks are arranged in descending order of the accuracy.}
    \label{fig:ReNARGNN}
\end{figure}

We integrate the framework with the Triplet-GMPNN processor~\cite{DBLP:conf/log/IbarzKPNBCDBVRD22}. As shown in~\cref{fig:ReNARGNN}, ReNAR-linear achieves only marginal improvements on a subset of tasks compared to the baseline Triplet-GMPNN. This suggests that the reconstruction objective alone is insufficient when the encoder lacks the capacity to capture structural dependencies present in the input. In contrast, our ReNAR model, equipped with a GNN-based encoder and gated residual connections, demonstrates consistent performance gains across a wide range of algorithmic tasks.

These results highlight that a well-designed encoder is essential for effectively supporting both the prediction and reconstruction tasks. In comparison, simple linear encoders may lack the expressiveness required to meet the dual demands of representation learning in neural algorithmic reasoning.

\newpage

\section{Ablation: Impact of Scaling Factor $\lambda$ in ReNAR}\label{sec:abl2}

This section investigates the influence of the scaling factor $\lambda$ on various algorithmic tasks. In our experiments, ReNAR was evaluated using a range of scaling factors {0.0, 0.05, 0.1, 0.15, 0.20}, where $\lambda$ represents the contribution of the enhanced encoder discussed earlier.

\begin{figure*}[htb]
    \centering
    \includegraphics[width=0.9\linewidth]{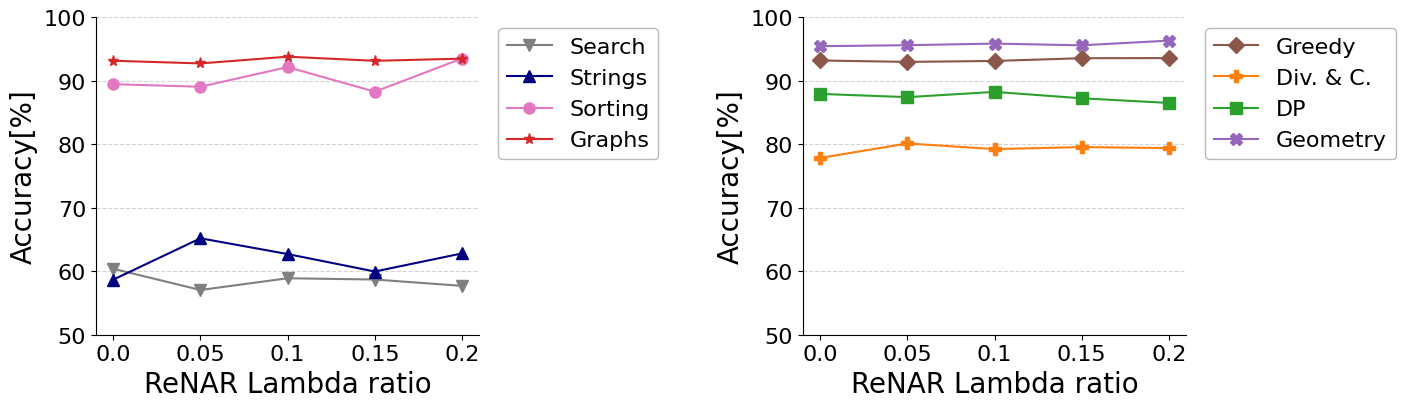}
    \caption{Ablation studies of scaling factor $\lambda$ in ReNAR.}
    \label{fig:ReNAR Lambda_Ratio}
\end{figure*}

As shown in ~\cref{fig:ReNAR Lambda_Ratio}, the effect of the scaling factor varies across different algorithmic categories. Overall, the impact of the scaling factor mirrors that of the mask ratios discussed previously. Specifically, graph-related tasks exhibit stable performance across all scaling factor values, while string-related tasks show more noticeable fluctuations. This difference is likely attributable to the inherent characteristics of the algorithms themselves.

\newpage
\section{Generality of the Framework Across Different Processors}\label{sec:abl3}
In this section, we present additional experimental results demonstrating the applicability of our proposed ReNAR and M-ReNAR framework to multiple processor architectures within the CLRS benchmark. The training hyperparameters are kept identical to those used for ReNAR and M-ReNAR in the main body. 

\begin{figure*}[htb]
    \centering
    \includegraphics[width=0.9\linewidth]{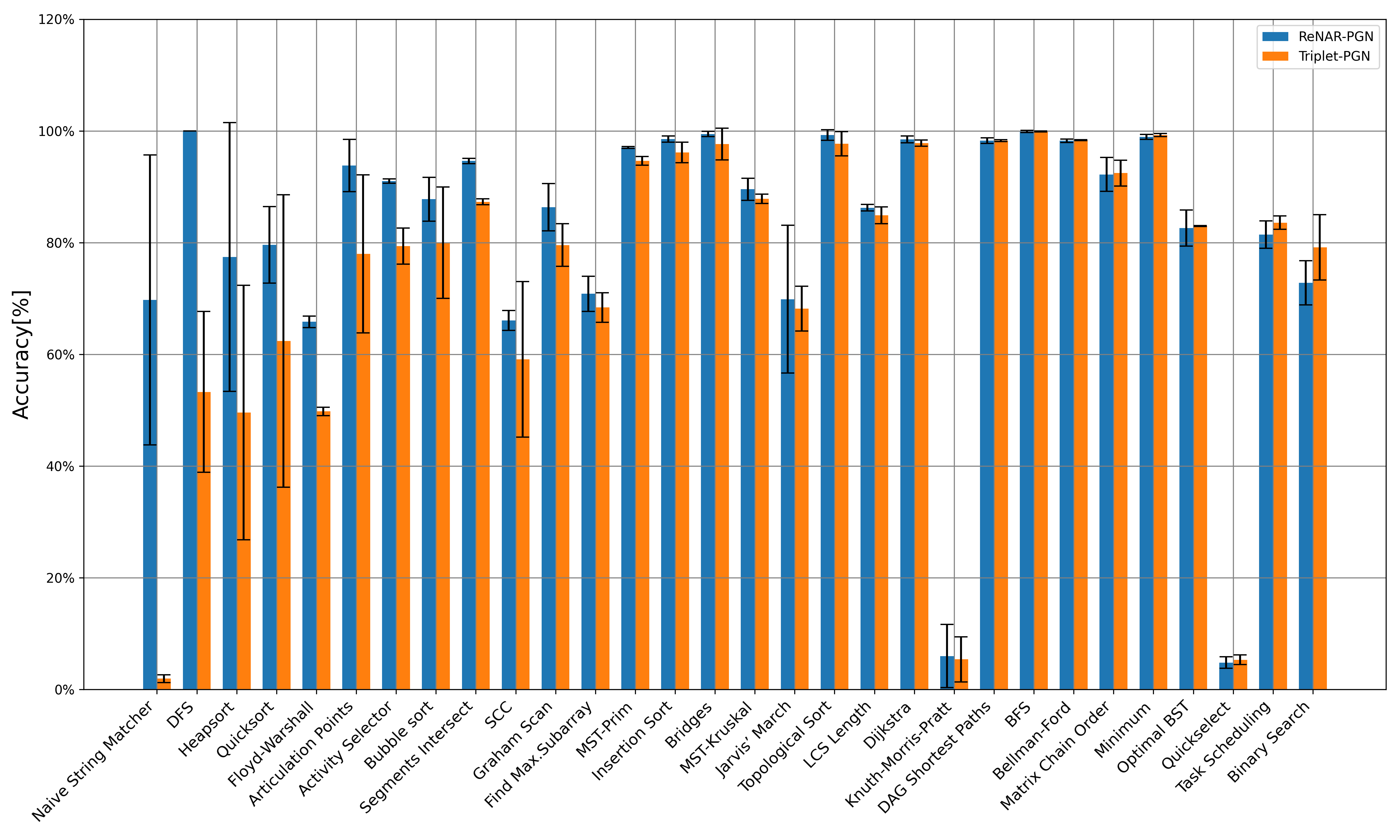}
    \caption{Comparison of the Triplet-PGN architecture’s performance before and after augmentation with ReNAR. The 30 algorithmic tasks are arranged in descending order of the accuracy.}
    \label{fig:ReNAR-Triplet-PGN}
\end{figure*}

\begin{figure*}[htb]
    \centering
    \includegraphics[width=0.9\linewidth]{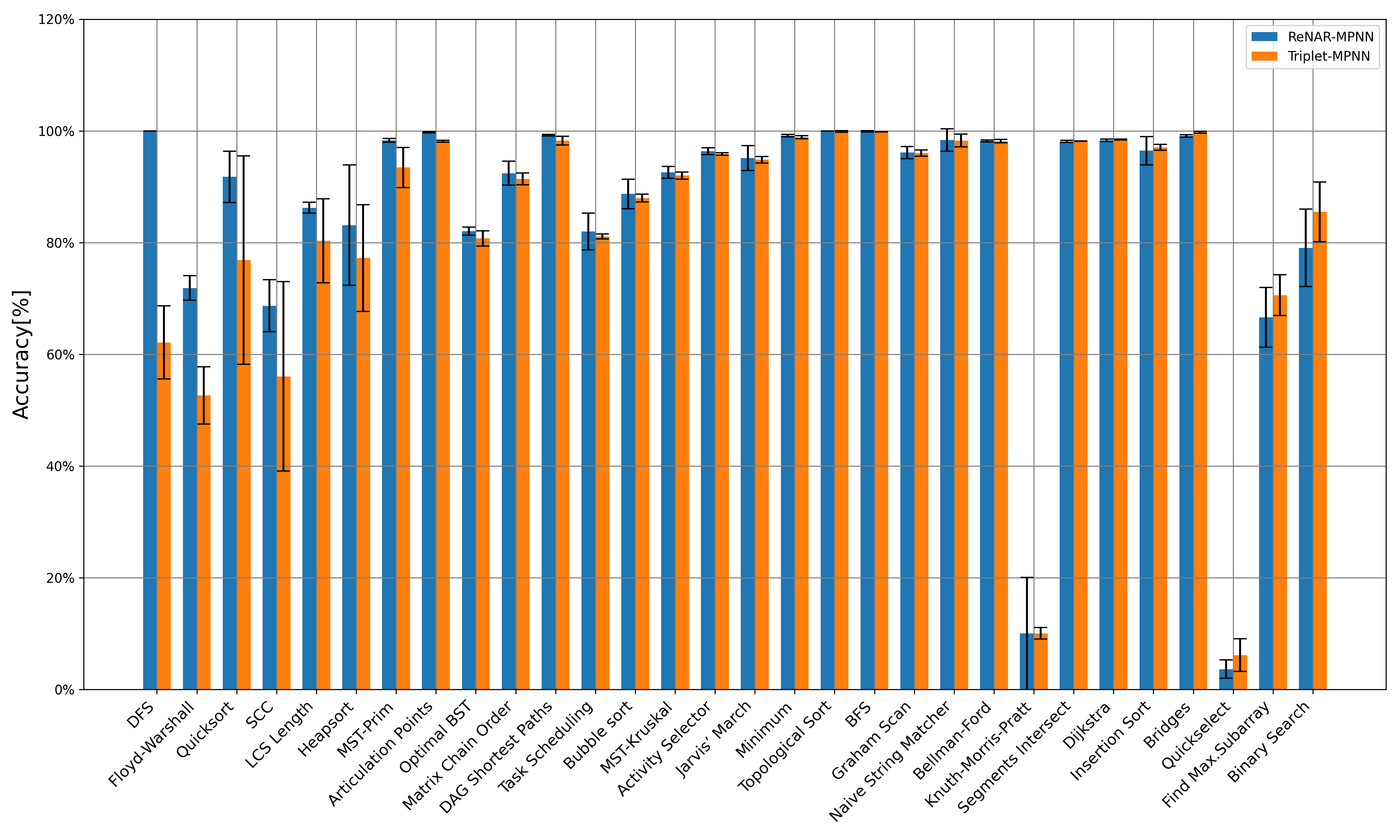}
    \caption{Comparison of the Triplet-MPNN architecture’s performance before and after augmentation with ReNAR. The 30 algorithmic tasks are arranged in descending order of the accuracy.}
    \label{fig:ReNAR-Triplet-MPNN}
\end{figure*}

\begin{figure*}[htb]
    \centering
    \includegraphics[width=0.9\linewidth]{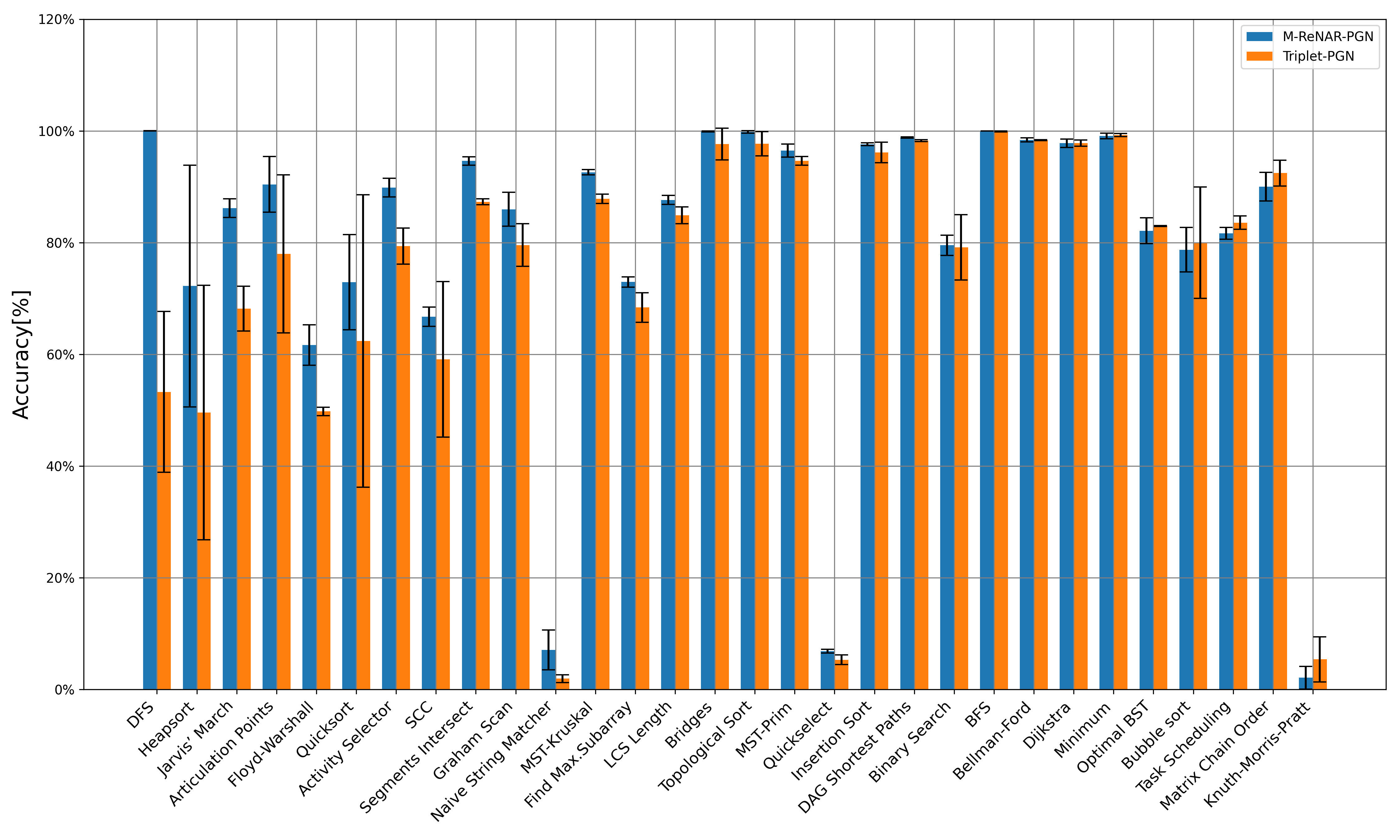}
    \caption{Comparison of the Triplet-PGN architecture’s performance before and after augmentation with M-ReNAR. The 30 algorithmic tasks are arranged in descending order of the accuracy.}
    \label{fig:Triplet-PGN}
\end{figure*}

\begin{figure*}[htb]
    \centering
    \includegraphics[width=0.9\linewidth]{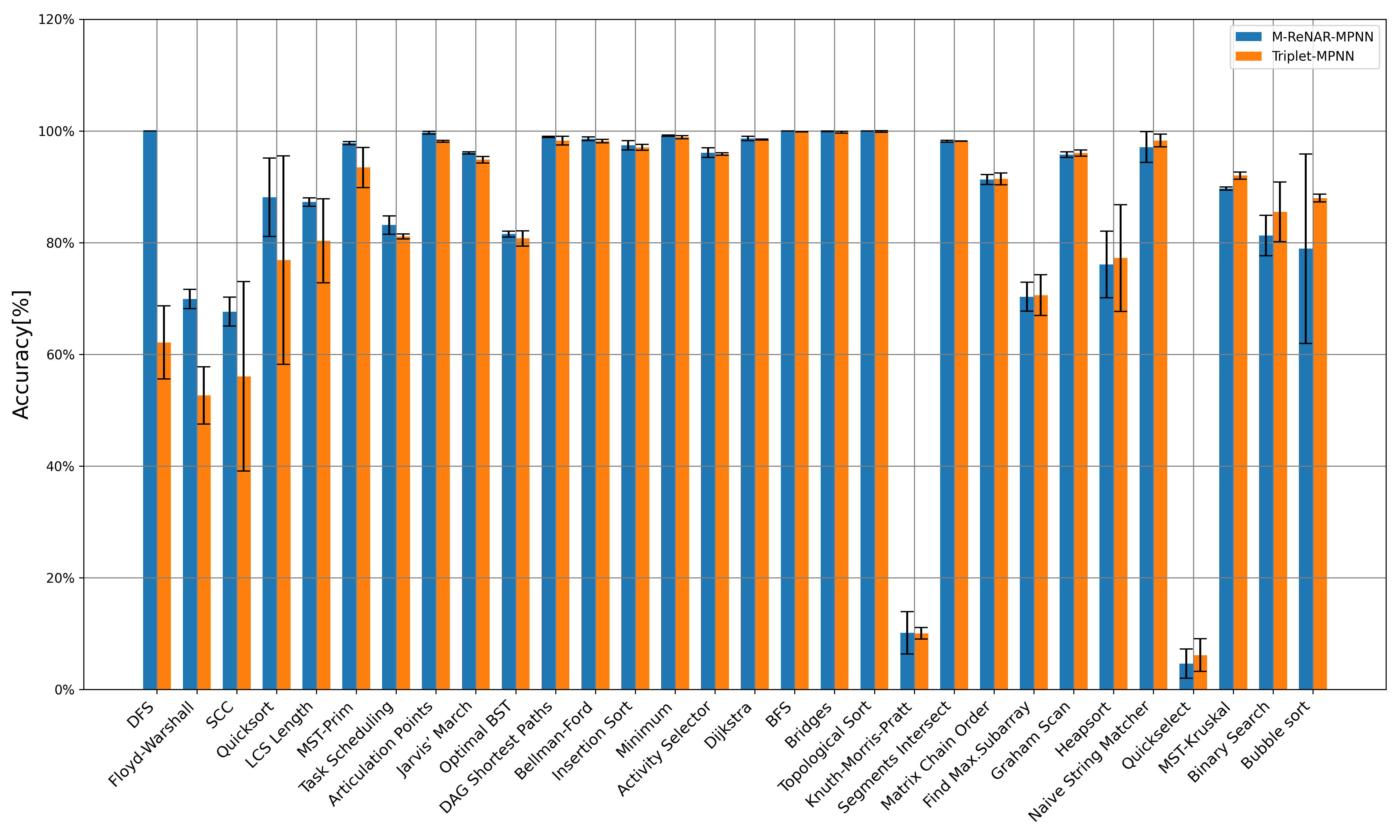}
    \caption{Comparison of the Triplet-MPNN architecture’s performance before and after augmentation with M-ReNAR. The 30 algorithmic tasks are arranged in descending order of the accuracy.}
    \label{fig:Triplet-MPNN}
\end{figure*}

To compare the performance of ReNAR and M-ReNAR, we evaluate both on the Triplet-PGN and Triplet-MPNN processors. The results for ReNAR are shown in~\Cref{fig:ReNAR-Triplet-PGN} and~\Cref{fig:ReNAR-Triplet-MPNN}, while those for M-ReNAR are presented in~\Cref{fig:Triplet-PGN} and~\Cref{fig:Triplet-MPNN}.

From~\Cref{fig:ReNAR-Triplet-PGN} and~\Cref{fig:ReNAR-Triplet-MPNN}, we observe that integrating our framework with different processors consistently leads to improvements in average accuracy. This indicates that the proposed ReNAR framework is broadly compatible and can serve as a general enhancement module across various neural algorithmic reasoning models.

\newpage


From ~\Cref{fig:Triplet-PGN} and~\Cref{fig:Triplet-MPNN}, we can see that integrating our M-ReNAR framework with different processors improves the average accuracy. These results suggest that M-ReNAR is broadly compatible and can function as a general enhancement module across diverse NAR architectures.


\end{document}